\documentclass[10pt,journal,compsoc]{IEEEtran}
\usepackage{hyperref}       
\hypersetup{colorlinks=true,urlcolor=blue}
\usepackage{url}            
\usepackage{booktabs}       
\usepackage{amsfonts}       
\usepackage{nicefrac}       
\usepackage{microtype}      
\usepackage{algorithm}
\usepackage{algorithmic}
\usepackage{graphicx}
\usepackage{epsfig}
\usepackage{amsmath}
\usepackage{amssymb}
\usepackage{amsthm}
\usepackage{subfigure}
\usepackage{booktabs} 
\usepackage{multirow}
\usepackage{etoolbox}
\usepackage{sail}
\usepackage{color}
\usepackage{footnote}
\usepackage{arydshln}
\usepackage{enumitem}
\usepackage{times}
\usepackage{ragged2e}
\usepackage{threeparttable}
\usepackage[export]{adjustbox}
\usepackage{bbm}

\usepackage{amsmath}
\usepackage{amssymb}
\def\hao{\textcolor{black}}

\usepackage{subfigure} 
\usepackage{longtable}

\usepackage{algorithm}
\usepackage{algorithmic}
\usepackage{hyperref}

\usepackage{amsfonts}

\usepackage{amsmath}

\def\etal{{\em et al.\/}\, }

\def\mM{{\mathcal M}}

\def\mP{{\mathcal P}}

\DeclareMathAlphabet\mathbfcal{OMS}{cmsy}{b}{n}

\def\0{{\bf 0}}
\def\1{{\bf 1}}

\def\bA{{\bf A}}

\def\bF{{\bf F}}

\def\bK{{\bf K}}

\def\bW{{\bf W}}
\def\bX{{\bf X}}
\def\bY{{\bf Y}}

\def\bx{{\bf x}}

\def\trsp{{\sf T}}

\def\bx{{\bf x}}
\def\bX{{\bf X}}

\def\bY{{\bf Y}}

\def\bW{{\bf W}}

\def\bK{{\bf K}}

\date{\today}

\usepackage{algorithm}

\ifCLASSOPTIONcompsoc
\usepackage[nocompress]{cite}
\else
\usepackage{cite}
\fi

\ifCLASSINFOpdf
\else
\fi

\begin{document}
	\title{Graph Convolutional Module for\\ Temporal Action Localization in Videos}
	
	\author{Runhao Zeng*, Wenbing Huang*, Mingkui Tan${}^\dagger$, Yu Rong, Peilin Zhao, Junzhou Huang, and Chuang Gan
		\IEEEcompsocitemizethanks{\IEEEcompsocthanksitem R. Zeng is with the School of Software Engineering, South China University of Technology and also with the Pazhou Laboratory, Guangzhou, China.\protect\\
			E-mail: runhaozeng.cs@gmail.com

			\IEEEcompsocthanksitem W. Huang is with the Department of Computer Science and Technology, Tsinghua University, State Key Lab. of Intelligent
			Technology and Systems, Tsinghua National Lab. for Information Science and Technology (TNList)\protect\\
			E-mail: hwenbing@126.com
			
			\IEEEcompsocthanksitem M. Tan is with the School of Software Engineering, South China University of Technology and also with the Key Laboratory of Big Data and Intelligent Robot, South China University of Technology, Ministry of Education. \protect\\
			E-mail: mingkuitan@gmail.com

			\IEEEcompsocthanksitem Y. Rong, P. Zhao and J. Huang are with the Tencent AI Laboratory, Shenzhen, China\protect\\
			E-mail: yu.rong@hotmail.com; peilinzhao@hotmail.com; jzhuang@uta.edu
			
			\IEEEcompsocthanksitem C. Gan is with the MIT-IBM Watson AI Lab, Cambridge, MA 02142 USA\protect\\
			E-mail: ganchuang1990@gmail.com
			
			\IEEEcompsocthanksitem  ${}^*$Authors contributed equally. ${}^\dagger$Corresponding author.
			}
			
		\thanks{Manuscript received April 19, 2005; revised August 26, 2015.}}

	\markboth{Journal of \LaTeX\ Class Files,~Vol.~14, No.~8, August~2015}%
	{Shell \MakeLowercase{\textit{et al.}}: Bare Demo of IEEEtran.cls for Computer Society Journals}

	\IEEEtitleabstractindextext{%
		\begin{abstract}
			\justifying
			Temporal action localization, which requires a machine to recognize the location as well as the category of action instances in videos, has long been researched in computer vision. The main challenge of temporal action localization lies in that videos  are usually long and untrimmed with diverse action contents involved. Existing state-of-the-art action localization methods divide each video into multiple action units (\emph{i.e.}, proposals in two-stage methods and segments in one-stage methods) and then perform action recognition/regression on each of them individually, without explicitly exploiting their relations during learning. 
			In this paper, we claim that the relations between action units play an important role in action localization, and a more powerful action detector should not only capture the local content of each action unit but also allow a wider field of view on the context related to it. To this end, we propose a general graph convolutional module (GCM) that can be easily plugged into existing action localization methods, including two-stage and one-stage paradigms. 
			To be specific, we first construct a graph, where each action unit is represented as a node and their relations between two action units as an edge. Here, we use two types of relations, one for capturing the temporal connections between different action units, and the other one for characterizing their semantic relationship. Particularly for the temporal connections in two-stage methods, we further explore two different kinds of edges, one connecting the overlapping action units and the other one connecting surrounding but disjointed units. Upon the graph we built, we then apply graph convolutional networks (GCNs) to model the relations among different action units, which is able to learn more informative representations to enhance action localization.
			Experimental results show that our GCM consistently improves the performance of existing action localization methods, including two-stage methods (\eg, CBR~\cite{gao2017cascaded} and R-C3D~\cite{xu2017r}) and one-stage methods (\eg, D-SSAD~\cite{huang2019decoupling}), verifying the generality and effectiveness of our GCM.
			Moreover, with the aid of GCM, our approach significantly outperforms the state-of-the-art on THUMOS14 (50.9\% versus 42.8\%). Augmentation experiments on ActivityNet also verify the efficacy of modeling the relationships between action units. 
		\end{abstract}
		
		\begin{IEEEkeywords}
			Temporal Action Localization, Graph Convolutional Networks, Video Analysis.
	\end{IEEEkeywords}}

	\maketitle

	\IEEEdisplaynontitleabstractindextext
	\IEEEpeerreviewmaketitle
	\IEEEraisesectionheading{\section{Introduction}\label{Sec:introduction}}
	
\begin{figure}[t]
		\centering
		\includegraphics[width=\linewidth]{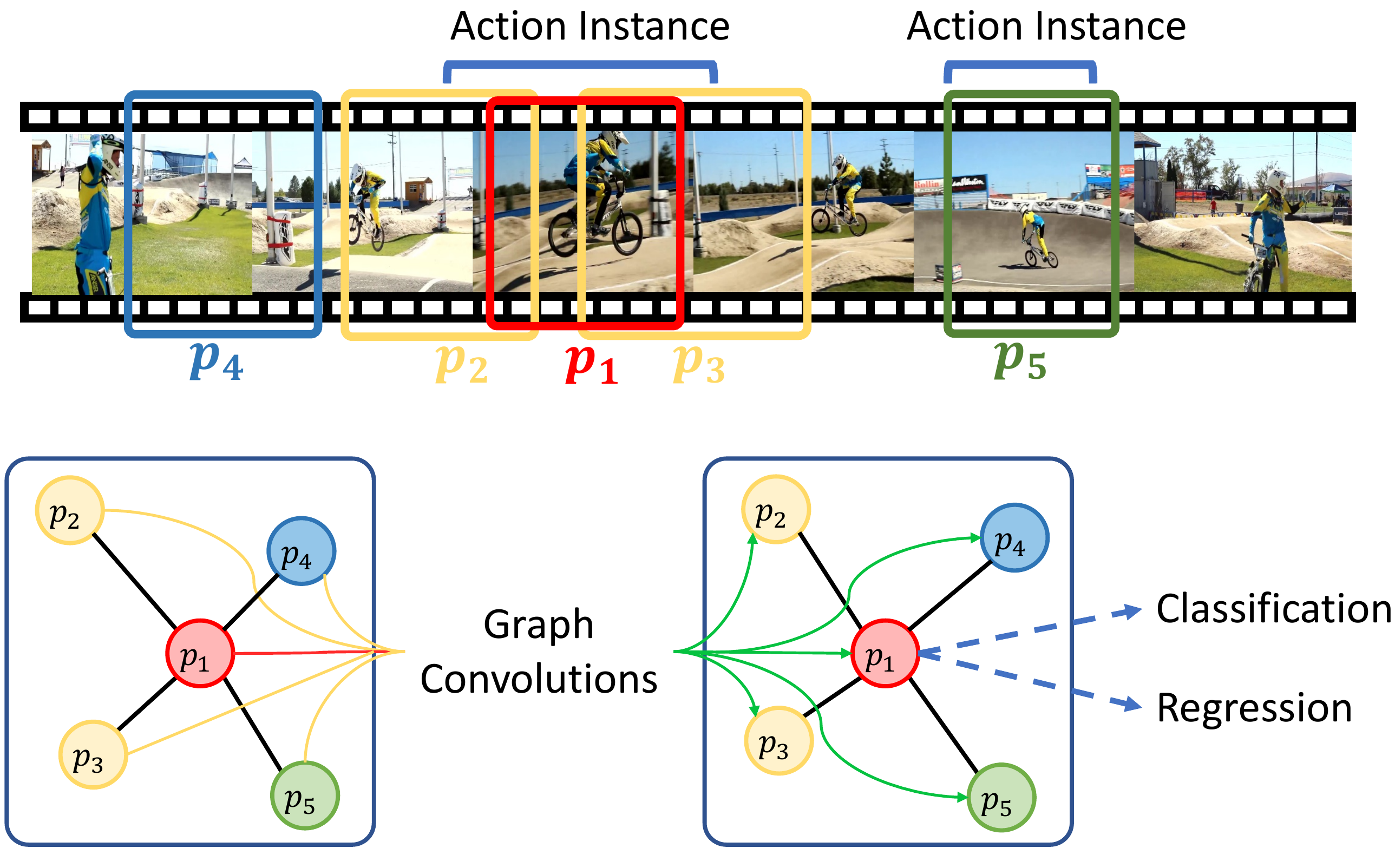}
		\caption{Schematic depiction of our approach. We apply graph convolutional networks to model the interactions between action units and boost the temporal action localization performance.}
		\label{Fig:simple}
	\end{figure}

	\IEEEPARstart{U}{nderstanding} human actions from raw videos is a long-standing research goal of computer vision, owing to its various applications in security surveillance, human behavior analysis and many other areas~\cite{simonyan2014two,tran2015learning,fan2018end,wang2016temporal}. Joining the success of deep learning, video-based action classification~\cite{wang2016temporal,carreira2017quo,tran2015learning} has exhibited fruitful progress in recent years. Nevertheless, this task assumes a tacit approval of addressing videos that are trimmed and short, which limits its practical potential. Temporal action localization, in contrast, targets on untrimmed and long videos to localize the start and end times of every action instance of interest as well as to predict the corresponding label. Taking the sports video in Figure~\ref{Fig:simple} as an example, the detector should determine where the action event is occurring and identify which class the event belongs to. The lower restriction in video collection and preprocessing makes temporal action localization a more compelling yet challenging task in video analytics.


	A variety of studies have been performed on temporal action localization over the last few years~\cite{chao2018rethinking,gao2017cascaded,lin2017single,shou2017cdc,shou2016temporal,buch2017end,huang2019decoupling,alwassel2018action,gleason2019proposal,zhao2017temporal}. In general, existing methods are categorized into two types: the two-stage paradigm~\cite{shou2016temporal,zhao2017temporal,chao2018rethinking,gao2017cascaded} and the one-stage paradigm~\cite{lin2017single,huang2019decoupling,buch2017end}. For the two-stage methods, they first generate a set of action proposals and then individually perform classification and temporal boundary regression on each proposal.
	In terms of one-stage methods, they divide each video into segments of equal number and then predict the labels and boundary offsets of the anchors mounted to each segment. Despite their difference in whether they use external proposals or not, these two paradigms share the similar spirit of independently conducting classification/regression on each action unit---a general concept corresponds to the proposal in two-stage methods and the segment in one-stage methods. Processing each action unit separately, however, will inevitably neglect the relations in-between and potentially lose critical cues for action localization. 
	For example, the adjacent action units around the target unit can provide temporal context for localizing its temporal boundary. Meanwhile, two distant action units might also provide indicative hints of action recognition to each other if they are semantically similar.
	
	Based upon the intuition above, this paper investigates the relationships between action units from two perspectives, namely the \emph{temporal relationship} and the \emph{semantic relationship}.
	To illustrate this, we revisit the example in Figure~\ref{Fig:simple}, where we have generated five action units.
	\textbf{1)} \emph{Temporal relationship}: the action units $\Mat{p}_1$, $\Mat{p}_2$ and $\Mat{p}_3$ overlapping with each other describe different parts of the same action instance (\ie, the start period, main body and end period).
	Conventional action localization methods perform prediction on $\Mat{p}_1$ by using its feature alone, which we think is insufficient to deliver complete knowledge.
	If we additionally consider the features of $\Mat{p}_2$ and $\Mat{p}_3$, we will obtain more contextual information around $\Mat{p}_1$, which is advantageous especially for the temporal boundary regression of $\Mat{p}_1$.
	On the other hand, $\Mat{p}_4$ describes the background (\ie, the sport field), and its content is also helpful in identifying the action label of $\Mat{p}_1$, since what is happening on the sports field is likely to be sports action (\eg, ``riding bicycle'') but not the action that occurs elsewhere  (\eg, ``kissing''). In other words, the classification of $\Mat{p}_1$ can be partly guided by the content of $\Mat{p}_4$ since they are temporally related even disjointed.
	\textbf{2)} \emph{Semantic relationship:} 
	$\Mat{p}_5$ is distant from $\Mat{p}_1$, but it describes the same action type as $\Mat{p}_1$ (``riding bicycle'') in a different view. We can acquire more complete information for predicting the action category of $\Mat{p}_1$ if we additionally leverage the content of $\Mat{p}_5$.
	
	To model the interactions between action units, one possible way is to employ the self-attention mechanism~\cite{vaswani2017attention}, as what has been conducted previously in language translation~\cite{vaswani2017attention} and object detection~\cite{hu2018relation}, to capture the pair-wise similarity between action units. 
	A self-attention module can affect an individual action unit by aggregating information from all other action units with the automatically learned aggregation weights.
	However, this method is computationally expensive as querying all action unit pairs has a quadratic complexity of the node number (note that each video can contain more than thousands of action units). In contrast, graph convolutional networks (GCNs), which generalize convolutions from grid-like data (\eg images) to non-grid structures (\eg social networks), have received increasing interest in the machine learning domain~\cite{kipf2017semi,yan2018spatial}. GCNs can affect each node by aggregating information from the adjacent nodes, and thus are very suitable for leveraging the relations between action units. More importantly, 
	unlike the self-attention strategy, 
	applying GCNs enables us to aggregate information from only the local neighborhoods for each action unit, and thus can help remarkably decrease the computational complexity.

	In this paper, we propose a general graph convolutional module (GCM) that can be easily plugged into existing action localization methods to exploit the relations between action units. 
	In this module, we first regard the action units as the nodes of a specific graph and represent their relations as edges. To construct the graph, we investigate  three kinds of edges between action units, including: \textbf{1)} the \emph{contextual edges} to incorporate the contextual information for each proposal instance (\eg, detecting $\Mat{p}_1$ by accessing $\Mat{p}_2$ and $\Mat{p}_3$ in Figure~\ref{Fig:simple}); \textbf{2)} the \emph{surrounding edges} to query knowledge from nearby but distinct action units (\eg, querying $\Mat{p}_4$ for $\Mat{p}_1$); \textbf{3)} the \emph{semantic edges} to
	involve the content of the semantically similar units for enhanced action recognition (\eg, recognizing $\Mat{p}_1$ by considering $\Mat{p}_5$). 
	Then, we perform graph convolutions on the constructed graph. Although the information is aggregated from local neighbors in each layer,
	message passing between distant nodes is still possible if the depth of the GCNs increases.
	Moreover, to avoid the overwhelming computational cost, we further devise a sampling strategy to train the GCNs efficiently while still preserving the desired detection performance. We evaluate our proposed method by incorporating GCM with existing action localization methods on two popular benchmarks for temporal action detection, \ie, THUMOS14~\cite{jiang2014thumos} and AcitivityNet1.3~\cite{caba2015activitynet}.
	
	In summary, our contributions are as follows:
	\begin{itemize}
		\item To the best of our knowledge, we are the first to exploit the relationships between action units for temporal action localization in videos. 
		\item To model the interactions between action units, we propose a general graph convolutional  module (GCM) to construct a graph of action units by establishing the edges based on our valuable observations and then apply GCNs for message aggregation among action units. Our GCM can be plugged  into existing two-stage and one-stage methods. 
		\item Experimental results show that GCM consistently improves the performance of SSN~\cite{zhao2017temporal}, R-C3D~\cite{xu2017r}, CBR~\cite{gao2017cascaded} and D-SSAD~\cite{huang2019decoupling} on two benchmarks, demonstrating the generality and effectiveness of our proposed GCM. On THUMOS14 especially, our method obtains a mAP of $50.9\%$ when $tIoU=0.5$, which significantly outperforms the state-of-the-art, \ie, $42.8\%$ by~\cite{chao2018rethinking}. Augmentation experiments on ActivityNet also verify the efficacy of modeling action proposal relationships.
	\end{itemize}
	
	This paper extends our preliminary version~\cite{zeng2019graph} that was published in ICCV 2019 in the following several aspects. 
	\textbf{1)} We integrate graph construction and graph convolution into a general graph convolutional module (GCM) so that the proposed module can be plugged into any of the two-stage temporal action localization methods (\eg, SSN, R-C3D and CBR) and the one-stage methods (\eg, D-SSAD). 
	\textbf{2)} In addition to the temporal relationships leveraged in our ICCV paper, we further explore semantic relationships to learn more discriminative representations. Experimental results reveal that the semantic relationships provide more valuable information for action recognition.
	\textbf{3)} We conduct more ablation studies (\eg, analysis of semantic edges, runtime comparison with the baseline methods, and comparisons for one-stage methods) to verify the  effectiveness and efficiency of the proposed method. \textbf{4)} We achieve clearly better action localization results over our ICCV version on THUMOS14 (50.9\% vs. 49.1\%) and ActivityNet 1.3 (31.45\% vs. 31.11\%).

	\section{Related work}\label{Sec:related}
	\textbf{Temporal action localization.} Recently, great progress has been achieved in deep learning~\cite{carreira2017quo,tran2015learning,wang2016temporal}, which facilitates the development of temporal action localization. Approaches on this task can be grouped into three categories: (1)  methods performing frame or segment-level classification, which requires a post-processing step (\eg, smoothing and merging) to obtain the temporal boundaries of the action instances~\cite{shou2017cdc,montes2016temporal,piergiovanni2019temporal}; (2) approaches employing a two-stage framework similar to the two-stage object detection methods in images. They often involve proposal generation, proposal classification and boundary refinement~\cite{shou2016temporal,xu2017r,zhao2017temporal}; (3) methods that integrate proposal generation and classification (and/or boundary regression) into end-to-end architectures, which are often called one-stage action localization methods~\cite{yeung2016end,buch2017end,lin2017single}. 
	
	Our work can be used to help both two-stage and one-stage action localization paradigms, where each video is divided into multiple action units and each action unit is processed individually. 
	Following the two-stage paradigm, Shou~\etal\cite{shou2016temporal} proposed generating a set of proposal candidates from sliding windows and classifying them by using deep neural networks.
	Xu~\etal\cite{xu2017r} exploited the 3D convolutional networks and proposed a framework inspired by Faster R-CNN~\cite{ren2015faster}. Following the one-stage paradigm, Lin~\etal\cite{lin2017single} divided the video into segments and used convolutional layers to obtain video features, which were further processed by an anchor layer for temporal action localization. Huang~\etal\cite{huang2019decoupling} decoupled the localization and classification in a one-stage scheme.
	However, the above methods neglect the contextual information of action units. To address this issue, some attempts have been developed to incorporate the context to enhance the proposal feature~\cite{dai2017temporal, gao2017turn, gao2017cascaded, zhao2017temporal, chao2018rethinking}. They show encouraging improvements by extracting features on the extended receptive field (\ie, boundary) of the proposal.
	Despite their success, they all process each action unit individually. In contrast, our method considered the relations between action units.
	
	\noindent \textbf{Graph-based relation modeling.}
	Relation modeling has proven to be very helpful in many computer vision tasks like object detection~\cite{hu2018relation}, visual reasoning~\cite{chen2018iterative} and image classification~\cite{wang2018non}. For instance, the performance of object detection can be improved by considering the object relations since objects in an image are often highly correlated~\cite{hu2018relation}. Recently, Kipf~\etal\cite{kipf2017semi} proposed graph convolutional network (GCN) to define convolutions on non-grid structures. Due to its effectiveness in relation modeling, GCN has been widely applied to several research areas in computer vision, such as skeleton-based action recognition~\cite{yan2018spatial}, object detection~\cite{xu2019spatial} and video classification~\cite{wang2018video}. Wang \etal\cite{wang2018video} used a graph to represent the spatiotemporal relations between objects for the action classification task. Xu \etal\cite{xu2019spatial} constructed an object graph relying on the spatial configurations between objects for object detection. Our work considers both the temporal and semantic relations between action units for a more challenging temporal action localization task, where both action classification and localization are required. Recently, Xu \etal\cite{xu2020gtad} proposed a one-stage action localization method with a graph to exploit the relations between video segments. Our work is able to model the relations between action units (\ie, video segments or proposals) and is more general since it can be easily plugged into existing action localization methods, including two-stage and one-stage paradigms.
	
	\noindent \textbf{Graph sampling strategy.} 
	For real-world applications, the graph can be large and directly using GCNs is inefficient. Therefore, several attempts have been made for efficient training by virtue of the sampling strategy, such as the node-wise method SAGE~\cite{hamilton2017inductive}, layer-wise model FastGCN~\cite{chen2018fastgcn} and its layer-dependent variant AS-GCN~\cite{huang2018adaptive}. In this paper, considering the flexibility and implementability, we adopt the SAGE method as the sampling strategy in our framework.
	
	\begin{figure*}[!t]
		\centering
		\includegraphics[width=\linewidth]{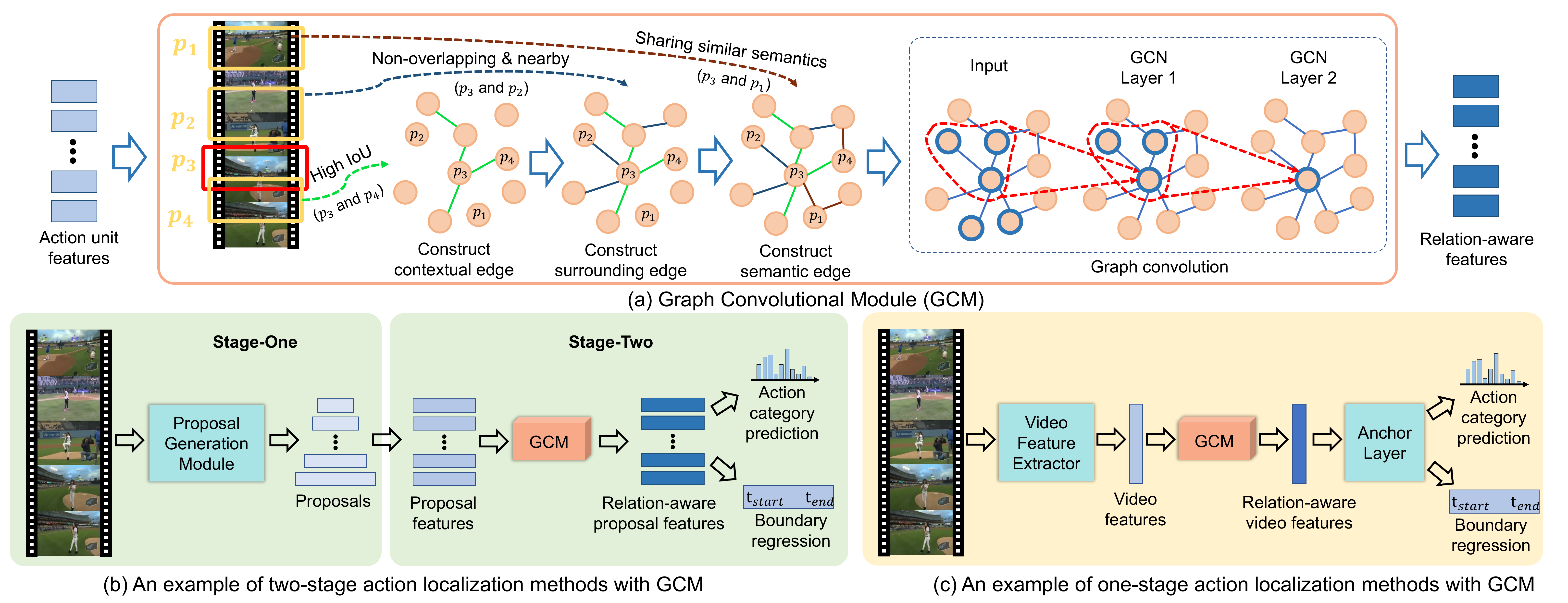}
		\caption{Schematic of our method. 
			(a) Given a set of action units (\eg, proposals in two-stage methods and segments in one-stage methods), our graph convolutional module (GCM) instantiates the nodes in the graph by each action unit. Then, we establish three kinds of edges among nodes to model the relations between action units and employ GCNs on the constructed graph. Lastly, our GCM module outputs relation-aware features. (b) For two-stage action localization methods, our GCM can be used in the second stage to enhance the proposal features, which are used for action classification and boundary regression. (c) For one-stage action localization methods, our GCM can be exploited to enhance the video features before the anchor layer. }
		\label{Fig:framework}
	\end{figure*}
	
	\section{Our Approach}\label{Sec:graph}
	
	\subsection{Notation and preliminaries} \label{Sec:notation}
	We denote an untrimmed video as $V=\{I_t\in\mathbb{R}^{H\times W\times 3}\}_{t=1}^T$, where $I_t$ denotes the frame at the time slot $t$ with height $H$ and width $W$.
	Within each video $V$,  let $\mP=\{\Mat{p}_i\mid\Mat{p}_i=(\bx_i, (t_{i,s}, t_{i,e}))\}_{i=1}^N$ be the action units of interest, where the action unit can be a proposal in two-stage action localization methods (\eg, SSN~\cite{zhao2017temporal}) or a video segment in one-stage methods (\eg, SSAD~\cite{lin2017single}).
	Let $t_{i,s}$ and $t_{i,e}$ be the start and end times of an action unit, respectively. In addition, given action unit $\Mat{p}_i$, let $\bx_i\in\mathcal{R}^d$ be the feature extracted by a certain feature extractor (\eg, the I3D network~\cite{carreira2017quo}) from frames between $I_{t_{i,s}}$ and $I_{t_{i,e}}$.
	
	Let $\mathcal{G}(\mathcal{V}, \mathcal{E})$ be a graph of $N$ nodes with nodes $v_i\in\mathcal{V}$ and edges $e_{ij}=(v_i, v_j)\in\mathcal{E}$. Furthermore, let $\bA\in\mathbb{R}^{N\times N}$ be the adjacency matrix  associated with $\mathcal{G}$.
	In this paper, we seek to exploit graphs $\mathcal{G}(\mathcal{P},\mathcal{E})$ on action units in $\mathcal{P}$ to better model the interactions between action units in videos. Here, each action unit is treated as a node, and the edges in $\mathcal{E}$ are used to represent the relations between nodes.
	
	\subsection{General scheme of our approach}
	
	
	We focus on solving the problem that existing temporal action localization methods neglect the relation between action units, which, however, is able to
	significantly improve the localization accuracy. Thus, we propose a general graph convolutional  module (GCM) that can be inserted into existing action localization methods in a plug-and-play manner. In particular, 
	GCM uses a graph $\mathcal{G}(\mathcal{P},\mathcal{E})$ to present the relations between action units and then applies GCN on the graph to exploit the relations and learn powerful representations for action units. The intuition is that when performing graph convolution, each node aggregates information from its neighborhoods. In this way, the feature of each action unit is enhanced by other action units, which helps eventually improve the detection performance. The schematic of our approach is shown in Figure~\ref{Fig:framework}.
	
	Without loss of generality, we assume the action units have been obtained beforehand by some methods (\eg, the TAG method in~\cite{zhao2017temporal}). Given the features of the action units $\{\bx_i\}_{i=1}^N$ and their initial temporal boundaries $\{(t_{i,s}, t_{i,e}))\}_{i=1}^N$, our GCM constructs a graph $\mathcal{G}$ according to the temporal and semantic relations between action units. Then, we apply a $K$-layer GCN in the GCM to exploit the relations and obtain the relation-aware  features $\bY$ of action units.
	For the $k$-th layer ($1\leq k\leq K$), the graph convolution is implemented by
	\begin{equation} 
	\label{Eq:gcn}
	\bX^{(k)} = \bA \bX^{(k-1)}\bW^{(k)}.
	\end{equation}
	Here, $\bA$ is the adjacency matrix, $\bW^{(k)}\in\mathbb{R}^{d_k\times d_k}$ is the parameter matrix to be learned, $\bX^{(k)}\in\mathbb{R}^{N \times d_k}$ are the hidden features for all action units at layer $k$, and $\bX^{(0)}\in\mathbb{R}^{N \times d}$ are the input features. 
	We apply an activation function (\ie, ReLU) after each convolution layer before the features are forwarded to the next layer. In addition, our experiments find it more effective by further combining the hidden features with the input features in the last layer, namely, 
	\begin{eqnarray}
	\label{Eq:layer-wise}
	\bY = \bX^{(K)} + \bX^{(0)},
	\end{eqnarray}
	where the summation is performed in an element-wise manner. The relation-aware action unit features $\bY$ are then used to jointly predict the action category $\hat{y}_i$ and temporal position $(\hat{t}_{i,s}, \hat{t}_{i,e})$ for each action unit $\Mat{p}_i$ by calculating
	\begin{eqnarray}
	\label{Eq:gcn-p}
	\{(\hat{y}_i, (\hat{t}_{i,s}, \hat{t}_{i,e}))\}_{i=1}^{N} = F(\bY), 
	\end{eqnarray}
	where $F$ denotes any action localization methods, such as SSN~\cite{zhao2017temporal}, R-C3D~\cite{xu2017r}, CBR~\cite{gao2017cascaded} and D-SSAD~\cite{huang2019decoupling}. 
	
	

	
	
	
	In the following sections, we aim to answer two questions: (1) how to construct a graph to represent the relations between action units, and (2) how to insert our GCM into the existing action localization methods, including the two-stage paradigm and one-stage paradigm.

	\subsection{Action unit graph construction}
	\label{Sec:construct}
	
	For the graph $\mathcal{G}(\mathcal{P},\mathcal{E})$ of each video, the nodes are instantiated as the action units, while the edges $\mathcal{E}$ between action units are demanded to be characterized specifically to better model the relations.
	\hao{One way for constructing edges is linking all action units with each other, which yet leads to overwhelming computations for going through all action unit pairs. It also incurs redundant or noisy information for action localization, as some unrelated action units should not be connected. In this paper, we devise a smarter approach by exploiting the temporal relevance/distance and the semantic relationships between action units instead. Specifically, we introduce three types of edges, the contextual edges, the surrounding edges and the semantic edges, respectively. }
	
	
	
	
	
	\subsubsection{Contextual edges}
	We establish an edge between action units $\Vec{p}_{i}$ and $\Vec{p}_{j}$ if $r(\Vec{p}_{i}, \Vec{p}_{j})> \theta_{ctx}$, where $\theta_{ctx}$ is a certain threshold.
	Here, $r(\Vec{p}_{i}, \Vec{p}_{j})$ represents the relevance between action units and is defined by the tIoU metric, \emph{i.e.}, 
	\begin{eqnarray}
	\label{Eq:relevance}
	r(\Vec{p}_{i}, \Vec{p}_{j}) = tIoU(\Vec{p}_{i}, \Vec{p}_{j})= \frac{I(\Vec{p}_{i}, \Vec{p}_{j})}{U(\Vec{p}_{i}, \Vec{p}_{j})},
	\end{eqnarray}
	where $I(\Vec{p}_{i}, \Vec{p}_{j})$ and $U(\Vec{p}_{i}, \Vec{p}_{j})$ compute the temporal intersection and union of the two action units, respectively. If we focus on the proposal $\Mat{p}_i$, establishing the edges by computing $r(\Vec{p}_{i}, \Vec{p}_{j})>\theta_{ctx}$ will select its neighborhoods as those that have high overlaps with it.  
	Obviously, the non-overlapping portions of the highly-overlapping neighborhoods can provide rich contextual information for $\Mat{p}_i$. As already demonstrated in~\cite{dai2017temporal,chao2018rethinking}, exploring the contextual information is of great help in refining the detection boundary and eventually increasing the detection accuracy. Here, by our contextual edges, all overlapping action units automatically share the contextual information with each other, and this information is further processed by the graph convolution.

	\subsubsection{Surrounding edges}
	The contextual edges connect the overlapping action units that usually correspond to the same action instance. Actually, surrounding but disjointed action units (including the background items) can also be correlated, and the message passing among them will facilitate the detection of each other. 
	For example, in Figure~\ref{Fig:simple}, the background $\Vec{p}_{4}$ provides guidance on identifying the action class of action unit $\Vec{p}_{1}$ (\eg, more likely to be sports actions).
	To handle such kind of correlations, we first utilize $r(\Vec{p}_{i}, \Vec{p}_{j})=0$ to query the disjointed action units, and then compute the following distance 
	\begin{eqnarray}
	\label{Eq:distance}
	d(\Vec{p}_{i}, \Vec{p}_{j})=\frac{|c_{i}-c_{j}|}{U(\Vec{p}_{i}, \Vec{p}_{j})},
	\end{eqnarray}
	to add the edges between nearby action units if $d(\Vec{p}_{i}, \Vec{p}_{j}) < \theta_{sur}$, where $\theta_{sur}$ is a certain threshold. In Eq.~\eqref{Eq:distance}, $c_{i}$ (or $c_{j}$) represents the center coordinate of $\Mat{p}_i$ (or $\Mat{p}_j$). 
	As a complement to the contextual edges, the surrounding edges enable the message to pass across distinct action instances and thereby provide more temporal cues for the detection. 
	
	
	\subsubsection{Semantic edges}
	\hao{The above contextual and surrounding edges aim to exploit the temporal context for each action unit, which, however, still neglects the semantic information between action units. It is worth noting that one untrimmed video often contains multiple action instances (\eg, each video on THUMOS14 dataset ~\cite{jiang2014thumos} contains more than 15 action instances on average), and the instances in one video often belong to the same or semantically similar action category. 
		For example, the actions \emph{CricketBowling} and \emph{CricketShot} often occur in the same video on THUMOS14. Although their categories are different when performing action localization, it is intuitive that the semantics of \emph{CricketBowling} are helpful for recognizing \emph{CricketShot} from other actions (\eg, \emph{CliffDiving}).
		Therefore, the proposal that locates at a distance from an action but containing similar semantic content might provide indicative hints for detecting the action.}
	
	\hao{To exploit such semantic information for action localization, we add a semantic edge between the action units that share similar semantics. In particular, we first define an action unit set $\mathcal{S}_i$ for the $i$-th action unit as  
		\begin{equation}
		\mathcal{S}_i=\{\Vec{p}_{j}| r(\Vec{p}_{i}, \Vec{p}_{j})=0, j \in \mathcal{N}_l (i)\},
		\end{equation} 
		where $\mathcal{N}_l (i)$ is the index set of the $l$ nearest neighborhoods of proposal $\Vec{p}_{i}$ and $\mathcal{N}_l (i)$ is constructed in the feature space relying on the cosine similarity between action unit features $\bx_{i}$ and $\bx_{j}$. Then, we establish a semantic edge between $\Vec{p}_{i}$ and the action units in $\mathcal{S}_i$. Note that the action unit feature $\bx_{i}$ can be the high-level appearance or motion feature containing rich semantic information. In other words, the action units sharing similar appearance (\eg, some similar places) or motions (\eg, the same action performed by different actors) can be used to help the recognition of action units. To summarize, the edge $e_{ij}$ between nodes $\Vec{p}_{i}$ and $\Vec{p}_{j}$ can be formulated as}	\begin{equation}
	\label{Eq:edge}
	e_{ij} =
	\begin{cases}
	1, & \mathrm{if}\quad r(\Vec{p}_{i}, \Vec{p}_{j})> \theta_{ctx}; \\
	1, &\mathrm{if}\quad r(\Vec{p}_{i}, \Vec{p}_{j})=0, d(\Vec{p}_{i}, \Vec{p}_{j}) < \theta_{sur}; \\
	1, &\mathrm{if}\quad r(\Vec{p}_{i}, \Vec{p}_{j})=0, j \in \mathcal{N}_l (i); \\
	0, & else.
	\end{cases}
	\end{equation}
	
	\subsubsection{Adjacency matrix}
	In Eq.~\eqref{Eq:gcn}, we need to compute the adjacency matrix $\bA$.
	Here, we design the adjacency matrix by
	assigning specific weights to edges. For example, we can apply the cosine similarity to estimate the weights of edge $e_{ij}$ by
	\begin{equation}
	\label{Eq:adj}
	\bA_{ij} =
	\begin{cases}
	\frac{\bx_{i}^{\trsp}\bx_{j}}{\|\bx_{i}\|_{2}\cdot\|\bx_{j}\|_{2}}, &  e_{ij}=1; \\
	0, & e_{ij}=0.
	\end{cases}
	\end{equation}
	In the above computation, we compute $\bA_{ij}$ relying on the feature vector $\bx_{i}$. We can also map the feature vectors 
	into an embedding space using a learnable linear mapping function as in \cite{wang2018non}
	before the cosine computation. We leave the discussion in our experiments. 
	
	
	\subsection{GCM for two-stage action localization methods}\label{Sec:module}
	
	Due to the residual nature of GCM (see Eq.~\eqref{Eq:layer-wise}), the proposed
	GCM can be easily plugged into existing two-stage action localization methods, which typically involve the following steps:
	\textbf{Step 1:} generates a set of proposal candidates, which may contain action instances; \textbf{Step 2:} uses some certain feature extractors, which can be off-the-shelf~\cite{gao2017cascaded} or trained in an end-to-end manner~\cite{xu2017r}, to obtain the proposal features; \textbf{Step 3:} processes the proposal features using an action classifier and a boundary regressor, which are often implemented as fully-connected layers; \textbf{Step 4:} performs duplicate removal, which is usually achieved by using non-maximum suppression (NMS).
	
	In this paper, our proposed GCM is used between Step 2 and Step 3. 
	Given a set of proposals, our GCM first constructs a proposal graph according to Equation~\eqref{Eq:edge}. Then, the relation-aware proposal features are obtained by performing graph convolution on the constructed graph via Equations~\eqref{Eq:gcn} and~\eqref{Eq:layer-wise}. 
	Joining the previous work SSN~\cite{zhao2017temporal}, we find that it is beneficial to predict the action label and temporal boundary separately by virtue of two GCMs---one conducted on the original proposal features $\bx_i$ and the other one on the extended proposal features $\bx'_{i}$. The first GCM is formulated as 
	\begin{equation}
	\label{Eq:1gcn}
	\{\hat{y}_i\}_{i=1}^{N} =  \mathrm{softmax}(\mathrm{FC}_1(\mathrm{GCM}_1(\{\bx_i\}_{i=1}^N))), \\
	\end{equation}
	where we apply a fully-connected (FC) layer with soft-max operation on top of $\mathrm{GCM}_1$ to predict the action label $\hat{y}_i$. The second GCM can be formulated as 
	\begin{eqnarray}
	\label{Eq:2gcn_1}
	\{(\hat{t}_{i,s}, \hat{t}_{i,e})\}_{i=1}^{N} = \mathrm{FC}_2(\mathrm{GCM}_2(\{\bx'_i\}_{i=1}^N)), \\
	\label{Eq:2gcn_2}
	\{\hat{e}_i\}_{i=1}^{N} = \mathrm{FC}_3(\mathrm{GCM}_2(\{\bx'_i\}_{i=1}^N)),
	\end{eqnarray}
	where the graph structure $\mathcal{G}(\mathcal{P},\mathcal{E})$ is the same as that in Eq.~\eqref{Eq:1gcn} but the input proposal feature is different. The extended feature $\bx'_{i}$ is attained by first extending the temporal boundary of $\Mat{p}_i$ with $\frac{1}{2}$ of its length on both the left and right sides and then extracting the feature within the extended boundary. Here, we adopt two FC layers on top of $\mathrm{GCM}_2$, one for predicting the boundary $(\hat{t}_{i,s}, \hat{t}_{i,e})$ and the other one for predicting the completeness score $\hat{c}_i$, which indicates whether the proposal is complete or not. It has been demonstrated by~\cite{zhao2017temporal} that, incomplete action units that have low tIoU with the ground-truths can have high classification scores, and thus it will make mistakes when using the classification score alone to rank the proposal for the mAP test; further applying the completeness score enables us to avoid this issue.
	
	For other two-stage action localization methods (\eg, CBR~\cite{gao2017cascaded}, R-C3D~\cite{xu2017r}) that do not rely on the two-stream pipeline such as SSN, we only insert one GCM into them. 
	Specifically, GCM takes the original proposal features $\bx_i$ as input and outputs the relation-aware features, which are further processed by two individual FC layers for predicting the action classification and boundary regression, respectively. Formally, the action localization process can be formulated as 
	\begin{eqnarray}
	\begin{aligned}
	\label{Eq:GCM2}
	\{(\hat{t}_{i,s}, \hat{t}_{i,e})\}_{i=1}^{N}& = \mathrm{FC}_4(\mathrm{GCM}_3(\{\bx_i\}_{i=1}^N)),\\
	\{\hat{y}_i\}_{i=1}^{N} &=  \mathrm{softmax}(\mathrm{FC}_5(\mathrm{GCM}_3(\{\bx_i\}_{i=1}^N))).
	\end{aligned}
	\end{eqnarray}
	where $\mathrm{FC}_*$ denotes the fully-connected (FC) layers, whose inputs are the same relation-aware features produced by GCM.

	\subsection{GCM for one-stage action localization methods}\label{Sec:one-stage}
	Our proposed GCM is a general module for exploiting the relationships between action units, which can be the segments in one-stage action localization methods, as discussed in Section~\ref{Sec:introduction}.
	
	Existing one-stage methods~\cite{huang2019decoupling,lin2017single} are inspired by the single-shot object detection methods in images~\cite{liu2016ssd}. A three-step pipeline is used in these methods, as summarized below.
	\textbf{Step 1:} evenly divides the input video into $T$ segments and extracts a $C$-dim feature vector for each segment, thus leading to a 1D feature map $\bF\in\mathbb{R}^{T \times C}$; \textbf{Step 2:} obtain 1D feature maps with multiple temporal scales (\ie, different temporal granularity) relying on $\bF$; \textbf{Step 3:} predict the action category and boundary offsets of the anchors mounted to each location on the 1D feature maps. For better readability, we call the feature vector at each location as a feature unit.
	
	Our proposed GCM is used between Step 2 and Step 3. 
	Although the boundaries of feature units are non-overlapping, we can incorporate our GCM to exploit the relations between feature units with a minor modification. 
	In particular, we only consider the surrounding and semantic edges to link the feature units and perform graph convolution to aggregate messages. The intuition is that the feature units can be regarded as a special case of proposals. Specifically, each feature unit corresponds to a segment in the videos with a certain duration, and these segments are non-overlapping. By adding the GCM to the 1D feature maps, we are able to exploit the relationship between the feature units in a 1D feature map. 
	It is worth mentioning that our module can be inserted one or multiple times throughout the network to model the feature relationships at different scales.

	\subsection{Training details}
	
	\subsubsection{Loss functions}
	\label{Sec:loss}
	Our proposed method not only predicts the action category and the completeness score (when inserting our GCM into SSN~\cite{zhao2017temporal}) of each proposal but also refines the temporal boundary of action units by location regression. To train our model, we define the following loss functions:
	
	\noindent \textbf{Classification Loss.} We define the training loss function for the action classifier as follows:
	\begin{equation}
	L_{cls}=\frac{1}{N}\sum_{i=1}^{N} L_{1}(y_i,\hat{y}_i),
	\end{equation}
	where $y_i$ and $\hat{y}_i$ are the ground truth and the prediction of the $i$-th action unit, respectively.
	We use the cross-entropy loss as $L_{1}$, and $N$ is the number of action units in a mini-batch.
	
	\noindent \textbf{Completeness Loss.}
	Given the predicted completeness score $\hat{e}_i$ and the ground truth $e_i$ of the $i$-th action unit, we use the following loss function to train the completeness predictor:
	\begin{equation}
	L_{com}=\frac{1}{N_{com}}\sum_{i=1}^{N} \mathbbm{1}_{com}^i L_{2}(e_i,\hat{e}_i),
	\end{equation}
	where we use hinge loss as $L_{2}$ and $N_{com}$ is the number of completeness training samples. $\mathbbm{1}_{com}^i$ is the indicator function, being 1 if $y_i\ge1$ (\ie, the action unit is not considered as part of the background) and 0 otherwise. 
	
	\noindent \textbf{Regression Loss.}
	We devise a set of location regressors $\{R_m\}_{m=1}^{N_{class}}$, each for an action category. For an action unit, we regress the boundary using the closest ground-truth instance as the target. Our method predicts the offset $\hat{o}_i=(\hat{o}_{i,c}, \hat{o}_{i,l})$ relative to the action unit , where $\hat{o}_{i,c}$ and $\hat{o}_{i,l}$ are the offset of center coordinate and length, respectively. The ground-truth offset is denoted as $o_i=(o_{i,c}, o_{i,l})$ and parameterized by:
	\begin{equation}
	\begin{aligned}
	o_{i,c}&=(c_i-c_{gt})/l_i, \\
	o_{i,l}&=log(l_i/l_{gt}), 
	\end{aligned}
	\end{equation}
	where $c_i$ and $l_i$ denote the original center coordinate and length of the action unit, respectively. $c_{gt}$ and $l_{gt}$ are the center coordinate and length of the closest ground truth, respectively. To train the regressor, we define the following loss function:
	\begin{equation}
	L_{reg}=\frac{1}{N_{reg}}\sum_{i=1}^{N} \mathbbm{1}_{reg}^i L_{3}(o_i, \hat{o}_i),
	\end{equation}
	where $N_{reg}$ is the number of regression training samples.
	$\mathbbm{1}_{reg}^i$ is the indicator function, being 1 if $y_i\ge1$ and $e_i=1$ (\ie, the proposal is a foreground sample) and 0 otherwise. We use the smooth-L1 loss as $L_{3}$ because it is less sensitive to outliers.
	
	\noindent \textbf{Multi-task Loss.} We train the whole model by using the following multi-task loss function:
	\begin{equation}\label{Eq:loss_total}
	L_{total}=L_{cls}+\lambda_{1}L_{com} +\lambda_{2}L_{reg},
	\end{equation}
	where $\lambda_{1}$ and $\lambda_{2}$ are hyper-parameters to trade-off these
	losses. We set $\lambda_{1}=\lambda_{2}=0.5$ in all the experiments and find that
	it works well across all of them. It is worth mentioning that we consider the completeness loss only when we plug our GCM into the SSN method~\cite{zhao2017temporal}.
	
	\begin{algorithm}[!bt]
		\caption{Training details of our method.}
		\begin{flushleft}
			\textbf{Input:} Action unit set $\mP=\{\Mat{p}_i\mid\Mat{p}_i=(\bx_i, (t_{i,s}, t_{i,e}))\}_{i=1}^N$;
			original action unit features $\{\bx_{i}^{(0)}\}_{i=1}^N$;
			extended action unit features $\{\bx'_{i}{}^{(0)}\}_{i=1}^N$;
			graph depth $K$; sampling size $N_s$
		\end{flushleft}
		\begin{flushleft}
			\textbf{Parameter:} Weight matrices $\bW^{(k)}$, $\forall k \in \{1,\dots,K\}$
		\end{flushleft}
		\begin{algorithmic}[1]
			\STATE instantiate the nodes by the action units $\Mat{p}_i$, $\forall \Mat{p}_i \in \mP$
			\STATE establish edges between nodes using Eq.~\eqref{Eq:edge}
			\STATE obtain an action unit graph $\mathcal{G}(\mathcal{P}, \mathcal{E})$
			\STATE calculate adjacent matrix using Eq.~\eqref{Eq:adj}
			\WHILE {not converges}
			\FOR {$k=1\dots K$}
			\FOR {$\Mat{p} \in \mathcal{P}$}
			\STATE sample $N_s$ neighborhoods of $\Mat{p}$
			\STATE aggregate information using Eq.~\eqref{Eq:gcn-sampling}
			\ENDFOR
			\ENDFOR
			\STATE predict action categories $\{\hat{y}_i\}_{i=1}^{N}$ using Eq.~\eqref{Eq:1gcn}
			\STATE perform boundary regression using Eq.~\eqref{Eq:2gcn_1}
			\STATE predict completeness score $\{\hat{c}_i\}_{i=1}^{N}$ using Eq.~\eqref{Eq:2gcn_2}
			\STATE compute $L_{total}$ using Eq.~\eqref{Eq:loss_total}
			\STATE update parameters via stochastic gradient descent
			\ENDWHILE
		\end{algorithmic}
		\label{Alg:forward}
	\end{algorithm}
	
	\subsubsection{Efficient training by sampling}
	\label{Sec:training}
	
	Typical action unit generation methods usually produce thousands of action units for each video. 
	Applying the aforementioned graph convolution (Eq.~\eqref{Eq:gcn}) on all action units demands many computations and large memory footprints. To accelerate the training of GCNs, several approaches~\cite{chen2018fastgcn,huang2018adaptive,hamilton2017inductive} have been proposed based on neighborhood sampling. Here, we adopt the SAGE method~\cite{hamilton2017inductive} in our method for its flexibility. 
	
	The SAGE method uniformly samples the fixed-size neighborhoods of each node layer-by-layer in a top-down passway. In other words, the nodes of the $(k-1)$-th layer are formulated as the sampled neighborhoods of the nodes in the $k$-th layer. After all nodes of all layers are sampled, SAGE performs the information aggregation in a bottom-up manner. Here, we specify the aggregation function to be a sampling form of Eq.~\eqref{Eq:gcn}, namely,
	\begin{equation} 
	\label{Eq:gcn-sampling}
	\bx_i^{(k)} = \left(\frac{1}{N_s} \sum_{j=1}^{N_s} \bA_{ij}\bx_j^{(k-1)} + \bx_i^{(k-1)}\right)\bW^{(k)},
	\end{equation}
	where node $j$ is sampled from the neighborhoods of node $i$, \ie, $j\in\mathcal{N}(i)$, and $N_s$ is the sampling size and is much less than the total number $N$. The summation in Eq.~\eqref{Eq:gcn-sampling} is further normalized by $N_s$, which empirically makes the training more stable. In addition, we also enforce the self-addition of its feature for node $i$ in Eq.~\eqref{Eq:gcn-sampling}.
	We do not perform any sampling when testing. For better readability, Algorithm~\ref{Alg:forward} depicts the algorithmic flow of our method.
	
	

	\section{Experiments}\label{Sec:exp}
	
	\subsection{Datasets}
	\textbf{THUMOS14 \cite{jiang2014thumos}} is a standard benchmark for action localization. 
	Its training set, known as the UCF-101 dataset, consists of 13320 videos. The validation, testing and background sets contain 1010, 1574 and 2500 untrimmed videos, respectively. 
	The temporal action localization task of THUMOS14, which contains videos over 20 hours from 20 sports classes, is very challenging
	since each video has more than 15 action instances and its 71\% frames are occupied by background items.
	Following the common setting in \cite{jiang2014thumos}, we apply 200 videos in the validation set for training and conduct evaluation on the 213 annotated videos from the testing set. 
	
	\noindent \textbf{ActivityNet \cite{caba2015activitynet}} is another popular benchmark for action localization on untrimmed videos. We evaluate our method on ActivityNet v1.3, which contains approximately 10K training videos and 5K validation videos corresponding to 200 different activities. Each video has an average of 1.65 action instances. Following the standard practice, we train our method on the training videos and test it on the validation videos. In our experiments, we contrast our method with the state-of-the-art methods on both THUMOS14 and ActivityNet v1.3, and perform ablation studies on THUMOS14.
	
	
	\subsection{Implementation details}
	
	\noindent \textbf{Evaluation metrics.} We use the mean average precision (mAP) as the evaluation metric. 
	A proposal is considered to be correct if its temporal IoU with the ground-truth instance is larger than a certain threshold and the predicted category is the same as this ground-truth instance.
	On THUMOS14, the tIOU thresholds are chosen from $\{0.1, 0.2, 0.3, 0.4, 0.5\}$; on ActivityNet v1.3, the IoU thresholds are from $\{0.5, 0.75, 0.95\}$, and we also report the average mAP of the IoU thresholds between 0.5 and 0.95 with the step of $0.05$.

	\noindent \textbf{Graph construction.}
	We construct the graph by fixing the values of $\theta_{ctx}$ as 0.7 and  $\theta_{sur}$ as 1 for both streams, which are selected by grid search. We adopt a 2-layer GCN since we observed no clear improvement with more than 2 layers but the model complexity is increased. For more efficiency, we choose $N_{s}=4$ in Eq.~\eqref{Eq:gcn-sampling} for neighborhood sampling unless otherwise specified.

	\noindent \textbf{Training.} 
	The initial learning rate is 0.001 for the RGB stream and 0.01 for the flow stream. 
	During training, the learning rates are divided by 10 every 15 epochs. The dropout ratio is 0.8.


	
	
	\noindent \textbf{Testing.} 
	We do not perform neighborhood sampling (\emph{i.e.}, Eq.~\eqref{Eq:gcn-sampling}) for testing. The predictions of the RGB and flow streams are fused using a ratio of 2:3.
	We multiply the classification score with the completeness score as the final score for calculating mAP.
	We then use non-maximum suppression (NMS) to obtain the final predicted temporal action units for each action class separately.
	We use 800 and 100 action units per video for computing mAPs on THUMOS14 and ActivityNet v1.3, respectively. 
	
	\noindent \textbf{Action units and features for two-stage methods.} The action units in two-stage methods refer to the action proposals.
	Our model is implemented under the two-stream strategy~\cite{simonyan2014two}: RGB frames and optical-flow fields.
	\textbf{1)} For SSN~\cite{zhao2017temporal}, we 
	first uniformly divide each input video into 64-frame RGB/optical-flow segments and 
	adopt a two-stream I3D model pre-trained on Kinetics~\cite{carreira2017quo}
	to obtain a 1024-dimensional feature vector for each segment.
	Upon the I3D features, we further apply max pooling across segments to obtain one 1024-dimensional feature vector for each proposal that is obtained by the BSN method~\cite{lin2018bsn}. Note that we do not finetune the parameters of the I3D model in our training phase. 
	In addition to the I3D features and BSN proposals, our ablation studies in Section~\ref{Sec:backbone} also explore other types of features (\eg, 2D features~\cite{lin2018bsn}) and proposals  (\eg, TAG action units~\cite{zhao2017temporal}).
	\textbf{2)} For CBR~\cite{gao2017cascaded}, we use the two-stream model~\cite{xiong2016cuhk} pre-trained on the ActivityNet v1.3 training set as the feature extractor. We use the proposals obtained from the proposal stage in~\cite{gao2017cascaded} to perform action localization.
	\textbf{3)} For R-C3D~\cite{xu2017r}, we use a 3D ConvNet modified from C3D~\cite{tran2015learning} to extract proposal features. We adopt the  proposals generated by the proposal subnet in~\cite{xu2017r} for a fair comparison.
	
	\noindent \textbf{Action units and features for one-stage methods.}
	The action units in one-stage methods refer to the video segments. We follow~\cite{huang2019decoupling} to use two-stream networks~\cite{simonyan2014two}
	pre-trained on Kinetics~\cite{carreira2017quo} to extract spatial and temporal feature representations for each video clip with length 512. We keep other settings (\eg, the learning rate, anchor settings) the same as those are used  in~\cite{huang2019decoupling} for fair comparisons.

	\subsection{Comparison with state-of-the-art results}
	\noindent \textbf{THUMOS14.}
	Our method is compared with the state-of-the-art methods in Table \ref{Tab:thumos}. GCM consistently boosts the performance of both two-stage methods (\eg, SSN~\cite{zhao2017temporal}, R-C3D~\cite{xu2017r}, CBR~\cite{gao2017cascaded}) and one-stage methods (\eg, D-SSAD~\cite{huang2019decoupling}) on THUMOS14, demonstrating the generality and effectiveness of our proposed GCM. With the aid of GCM, our method (\ie, SSN+GCM) reaches the highest mAP over all thresholds, implying that our method can recognize and localize actions much more accurately than any other method. 
	In particular, our method outperforms the previously best method (\emph{i.e.}, TAL-Net~\cite{chao2018rethinking}) by 8.1\% absolute improvement and the second-best result~\cite{gleason2019proposal} by more than 13.5\%, when $tIoU=0.5$. When using the proposals of higher quality (\ie, BMN proposals~\cite{lin2019bmn}), our method (\ie, SSN+GCM${}^\dagger$) lifts the mAP to 51.9\% when $tIoU=0.5$.
	
	\begin{table}[!tb]
		\centering
		\caption{Action localization results on THUMOS14, measured by mAP (\%) at different tIoU thresholds $\alpha$. (${}^\dagger$) indicates the method that uses BMN proposals~\cite{lin2019bmn}.}
		\setlength{\tabcolsep}{4pt}
		\begin{tabular}{clccccccc}
			\hline
			Paradigm & tIoU                         & 0.1           & 0.2           & 0.3           & 0.4           & 0.5                     \\ \hline
			\multirow{5}{*}{One-Stage} & 
			Yeung \etal \cite{yeung2016end}     & -          & -          & 36.0          & 26.4          & 17.1          \\ 
			&Lin \etal \cite{lin2017single}     & -          & -          & 43.0          & 35.0          & 24.6          \\ 
			&Buch \etal \cite{buch2017end}            & -             & -             & 45.7          & -             & 29.2                    \\ \cline{2-7}
			&Huang \etal \cite{huang2019decoupling}                   & 66.4 &64.7 & 59.8& 53.4& 43.2 \\  
			&Huang \etal + GCM       & 66.4 & 65.2 & 61.4 & 54.7 & 44.8    \\ \hline \hline
			\multirow{28}{*}{Two-Stage}&Wang \etal \cite{wang2014action}         & 18.2          & 17.0          & 14.0          & 11.7          & 8.3                     \\
			&Caba \etal \cite{caba2016fast}           & -             & -            & -             & -             & 13.5  \\
			&Escorcia \etal  \cite{escorcia2016daps}  & -             & -             & -             & -             & 13.9                   \\
			&Oneata \etal \cite{oneata2014lear}       & 36.6          & 33.6          & 27.0          & 20.8          & 14.4                  \\
			&Richard \etal \cite{richard2016temporal} & 39.7          & 35.7          & 30.0          & 23.2          & 15.2                 \\
			&Yeung \etal \cite{yeung2016end}          & 48.9          & 44.0          & 36.0          & 26.4          & 17.1              \\
			&Yuan \etal \cite{Yuan2017}       & 51.0          & 45.2          & 36.5          & 27.8          & 17.8                    \\
			&Yuan \etal \cite{yuan2016temporal}       & 51.4          & 42.6          & 33.6          & 26.1          & 18.8                   \\
			&Shou \etal \cite{shou2016temporal}       & 47.7          & 43.5          & 36.3          & 28.7          & 19.0                   \\
			&Hou \etal \cite{hou2017real}             & 51.3          & -             & 43.7          & -            & 22.0                      \\
			&Buch \etal  \cite{buch2017sst}           & -             & -             & 37.8          & -             & 23.0                    \\
			&Shou \etal \cite{shou2017cdc}            & -             & -            & 40.1          & 29.4          & 23.3                    \\
			&Dai \etal \cite{dai2017temporal}         & -             & -            & -             & 33.3          & 25.6                  \\
			&Gao \etal \cite{gao2017turn}             & 54.0          & 50.9          & 44.1          & 34.9          & 25.6                   \\
			&Huang \etal  
			\cite{huang2018sap}      & -          & -          & -          & -          & 27.7                   \\
			&Yang \etal \cite{yang2018exploring}       & -          & -          & 44.1          & 37.1          & 28.2                   \\
			&Zhao \etal \cite{zhao2017temporal}       & 66.0          & 59.4          & 51.9          & 41.0          & 29.8                   \\ 
			&Gao \etal \cite{gao2018ctap}        & -          & -          & -          & -          & 29.9                 \\	
			&Alwassel \etal \cite{alwassel2018action}       & -          & -          & 51.8          & 42.4          & 30.8                   \\
			&Lin \etal \cite{lin2018bsn}              & -             & -             & 53.5          & 45.0          & 36.9                  \\
			&Gleason \etal
			\cite{gleason2019proposal} 
			& 52.1         & 51.4          & 49.7          & 46.1          & 37.4                   \\	 
			&Chao \etal \cite{chao2018rethinking}     & 59.8          & 57.1          & 53.2          & 48.5          & 42.8          \\ \cline{2-7}
			&Xu \etal  \cite{xu2017r}                  & 54.5          & 51.5          & 44.8          & 35.6          & 28.9          \\ 
			&Xu \etal + GCM        & 56.3 & 53.5 & 47.0 & 37.9 & 30.9  \\ 
			&Gao \etal \cite{gao2017cascaded}         & 60.1          & 56.7          & 50.1          & 41.3          & 31.0    \\  
			&Gao \etal +  GCM       & 61.2 & 57.8 & 50.3 & 42.4 & 32.2    \\ 
			&Zhao \etal \cite{zhao2017temporal} (I3D)       & 69.7 & 67.5 & 64.6 & 58.3 & 49.3    \\ 
			&Zhao \etal +  GCM                          & 70.5 & 68.6 & 65.2 & 59.8 & 50.9  \\ 
			&Zhao \etal +  GCM${}^\dagger$                      
			& \textbf{72.5} & \textbf{70.9} & \textbf{66.5} & \textbf{60.8} & \textbf{51.9}  \\\hline
			
		\end{tabular}
		\label{Tab:thumos}
	\end{table}

	\noindent \textbf{ActivityNet v1.3.} We report the action localization results of various methods in Table \ref{Tab:anet}. Regarding the average mAP, our method (\ie, SSN+GCM) outperforms SSN~\cite{zhao2017temporal}, and CDC~\cite{shou2017cdc} by 2.83\% and 3.40\%, respectively. 
	We observe that BSN~\cite{lin2018bsn} and BMN~\cite{lin2019bmn} perform promisingly on this dataset. Note that these two methods were originally designed for generating class-agnostic proposals, and thus rely on external video-level action labels (from UntrimmedNet \cite{wang2017untrimmed}) for action localization.
	In contrast, our method is self-contained and is able to perform action localization without any external label. 
	
	Actually, our method can be modified to take external labels into account. 
		To achieve this, we replace the predicted action classes in Eq.~\eqref{Eq:1gcn} with the external action labels. Specifically, given an input video, we use UntrimmedNet to predict the top-2 video-level classes and assign these classes to all the proposals in this video. Thus, each proposal has two predicted action classes.
		To compute mAP, we follow \cite{lin2018bsn} to obtain the score of each proposal by calculating $s_{prop} = s_{gcm} * s_{bsn/bmn} * s_{unet}$,
		where $s_{gcm}$ is the proposal score predicted by our model (\ie, SSN+GCM),
		$s_{bsn/bmn}$ is the confidence score produced by BSN (or BMN)
		and $s_{unet}$ denotes the action score predicted by UntrimmedNet.
		As summarized in Table~\ref{Tab:anet},  our enhanced version (\ie, SSN*+GCM) consistently outperforms BSN and BMN when using the same proposals. Moreover, SSN*+GCM outperforms GTAD~\cite{xu2020gtad} even though GTAD uses additional video classification scores from~\cite{wang2017untrimmed}. These results further demonstrate the effectiveness of our method.
	
		\begin{table}[!t]
		\centering
		\caption{Action localization results on ActivityNet v1.3 (val), measured by mAP (\%) at different tIoU thresholds and the average mAP of IoU thresholds from 0.5 to 0.95. (*) indicates the method that uses the external video labels/scores from UntrimmedNet \cite{wang2017untrimmed}.}
		\begin{tabular}{lccc|c}
			\hline
			tIoU                         & 0.5           & 0.75          & 0.95           & Average              \\ \hline
			Singh \etal  \cite{singh2016untrimmed}        & 34.47        & -             & -               & -                 \\
			Wang \etal   \cite{wang2016uts}        & 43.65        & -             & -               & -                 \\
			Shou \etal \cite{shou2017cdc}         & 45.30             & 26.00             & 0.20          & 23.80                      \\
			Dai \etal \cite{dai2017temporal}        & 36.44             & 21.15             & 3.90             & -                        \\
			Xu \etal \cite{xu2017r}         & 26.80          & -          & -          & -                         \\
			Zhao \etal \cite{zhao2017temporal}        & 39.12 & 23.48          & 5.49           & 23.98                       \\ 
			Chao \etal  \cite{chao2018rethinking}          & 38.23          & 18.30          & 1.30          & 20.22         \\  
			Lin \etal  \cite{lin2018bsn} (BSN*)    & 46.45        & 29.96          & 8.02        & 30.03 \\ 
			Xu \etal \cite{xu2020gtad} (GTAD*)   & 50.36        & 34.60          & 9.02        & 34.09 \\ 
			Lin \etal  \cite{lin2019bmn}  (BMN*)    & 50.07        & 34.78          & 8.29        & 33.85 \\ 
			\hline
			SSN (BSN prop \cite{lin2018bsn})     & 38.59   & 24.53  & 4.57 & 24.37  \\ 	
			SSN  + GCM  (BSN prop \cite{lin2018bsn})             & 42.55         & 28.27 & 2.84 & 27.20  \\ 
			SSN* + GCM (BSN prop \cite{lin2018bsn})  & 47.92  & 32.91 & 4.16 & 31.45 \\
			SSN*  + GCM (BMN prop \cite{lin2019bmn})  & 51.03  & 35.17 & 7.44 & 34.24 \\
			\hline
		\end{tabular}
		\label{Tab:anet}
	\end{table}

	
		\begin{table}[!tb]
		\centering
		\caption{Ablation study of GCM on CBR and R-C3D, measured by mAP (\%) when tIoU=0.5 on THUMOS14.}
		\begin{tabular}{l|cc}
			\hline
			Setting                & mAP@IoU=0.5     & Gain            \\ \hline
			CBR~\cite{gao2017cascaded}        & 31.00      &  -     \\
			CBR + GCM    & \textbf{32.24}    & 1.24   \\ \hline 
			R-C3D~\cite{xu2017r}  & 28.90   & -        \\
			R-C3D + GCM   & \textbf{30.85}   & 1.95 \\ \hline
		\end{tabular}
		\label{Tab:cbr}
	\end{table} 

\begin{figure*}[!t]
	\centering
	\includegraphics[width=\linewidth]{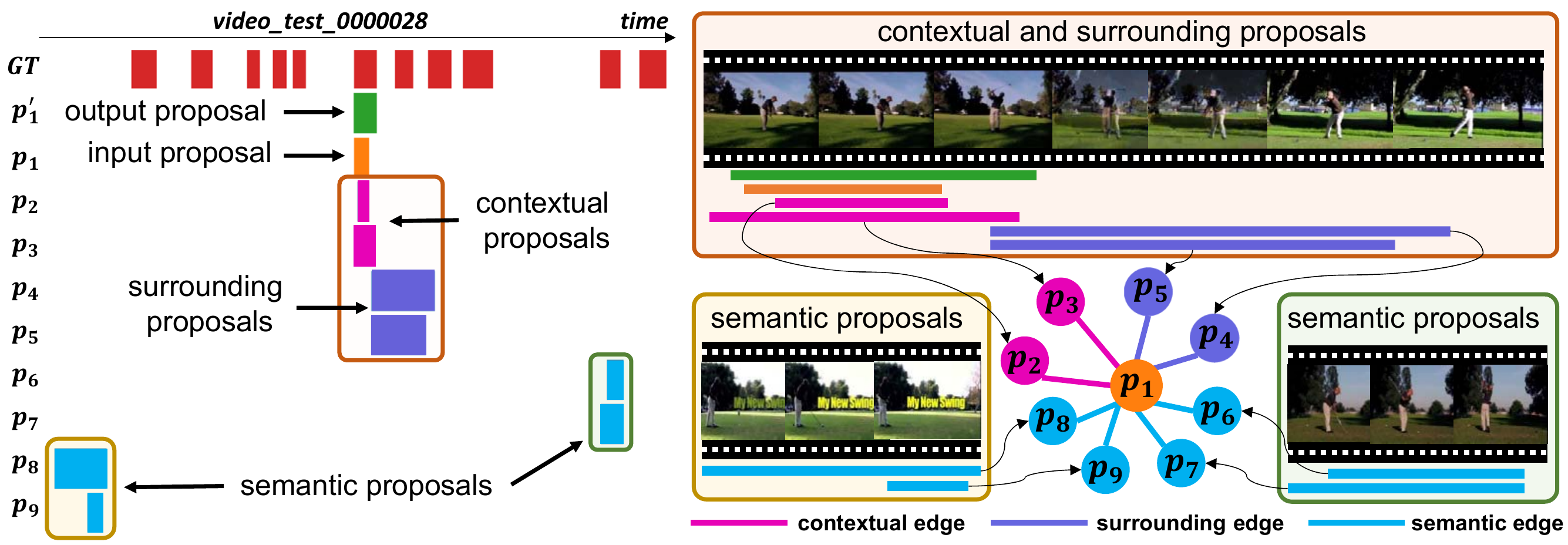}
    \caption{Visualization results of the graph constructed by our GCM on THUMOS14. The temporal boundary of the input proposal is not precise (\ie, some portions of the corresponding ground truth have not been detected). Our proposed GCM helps to aggregates contextual information from other proposals and lastly predicts the action category correctly and refines the temporal boundary of the input proposal precisely.
    }
    \label{Fig:graph}
    \end{figure*}
	
	\section{Ablation Results on Two-Stage Methods}
	In this section, we will perform complete and in-depth ablation studies to evaluate the impact of each component of our model. 
	
	\subsection{Effectiveness and generality of GCM}
	
	\hao{In this section, we incorporate our GCM into two popular two-stage action localization methods (\ie, CBR~\cite{gao2017cascaded} and R-C3D~\cite{xu2017r}) to validate the effectiveness and generality of GCM. In the following, we present the implementation details and the results.}
	
	\noindent \textbf{Cascaded Boundary Regression (CBR)~\cite{gao2017cascaded}}. \hao{The CBR method adopts a cascaded framework to iteratively regress the boundary of the action units. In the proposal stage, CBR uses a deep model to obtain the initial action units by refining the boundary of the sliding windows. In the detection stage, CBR uses another deep model to learn better representations of the action units. Last, these action units are forwarded to fully-connected layers for action classification and boundary regression. In our experiments, we insert our GCM in the detection stage. The outputs of the GCM are forwarded to the action classifier and regressor.
		For a fair comparison with CBR, we use two-stream features and unit-level offsets. As shown in Table~\ref{Tab:cbr}, our GCM helps to lift the action localization results over all IoU thresholds, demonstrating its effectiveness.}

	\noindent \textbf{Region Convolutional 3D Network (R-C3D)~\cite{xu2017r}}. \hao{Inspired by the faster-RCNN~\cite{ren2015faster} approach in object detection, Xu \etal proposed an end-to-end R-C3D network for activity detection. 
		The network encodes the frames with fully-convolutional 3D layers and then uses a proposal subnet to generate activity segments (\ie, action units). Last, they use a classification subnet to classify and refine the action units based on the RoI-pooled features. Our GCM takes the pooled features as input and enhances the features by constructing a graph and performing graph convolution. The outputs of the GCM are forwarded to the action classifier and regressor.
		We follow the same settings in ~\cite{xu2017r} for a fair comparison.  From Table~\ref{Tab:cbr}, the action localization performance of R-C3D is significantly improved with the help of our GCM. More critically, the performance gain on two action localization methods demonstrates the generality of our module.}

	\label{Sec:ablation}
	\subsection{How do proposal-proposal relations help?}
	\label{Sec:mlp}
	As illustrated in Section~\ref{Sec:module}, we apply two GCMs for action classification and boundary regression separately. 
	Here, we implement the baseline with a 2-layer multilayer-perceptron (MLP). The MLP baseline shares the same structure as GCM except that we remove the adjacent matrix $\bA$ in Eq.~\eqref{Eq:gcn}. Specifically, for the $k$-th layer, the propagation in Eq.~\eqref{Eq:gcn} becomes $\bX^{k}=\bX^{k-1}\bW^k$, where $\bW^k$ are the trainable parameters. 
	Without using $\bA$, MLP processes each proposal feature independently. By comparing the performance of MLP with GCN, we can justify the importance of message passing along action units. 
	To do so, we replace each GCM with an MLP and have the following variants of our model including: (1) \textbf{MLP$_1$ + GCM$_2$} where GCN$_1$ is replaced; (2) \textbf{GCM$_1$ + MLP$_2$} where GCM$_2$ is replaced; and (3) \textbf{MLP$_1$ + MLP$_2$} where both GCMs are replaced.
	Table \ref{Tab:twographs} shows that all these variants decrease the performance of our model, thus verifying the effectiveness of GCNs for both action classification and boundary regression. 
	Overall, our method significantly outperforms the MLP protocol (\ie \textbf{MLP$_1$ + MLP$_2$}), validating the importance of considering the relations between action units in temporal action localization. The MLP baseline is indeed a particular implementation of SSN~\cite{zhao2017temporal}. We compare the runtime between GCM and MLP baseline in Table~\ref{tab:runtime}. In detail, we train each model with 200 iterations on a Titan X GPU and report the average processing time per video per iteration (note that proposal generation and feature extraction are excluded for each model).
	It reads that GCM only incurs a relatively small additional runtime compared with the MLP baseline but is able to improve the performance significantly. 
	
	\noindent \textbf{Visualization of the constructed graph.} To understand how proposal-proposal relations help improve the action localization performance, we visualize an example of the graph constructed by our proposed method in Figure~\ref{Fig:graph}. Specifically, given an input proposal $\Mat{p}_1$, we choose $K=8$ proposals with the largest weights among all connected proposals. The temporal boundary of the input proposal $\Mat{p}_1$ is not precise (\ie, the ending period of the corresponding ground truth action instance has not been detected in $\Mat{p}_1$). The contextual and surrounding edges connect four proposals ($\Mat{p}_2$, $\Mat{p}_3$, $\Mat{p}_4$ and $\Mat{p}_5$) that can provide a wider receptive field for $\Mat{p}_1$ to detect the ending period of actions. Interestingly, the semantic edges connect not only two proposals ($\Mat{p}_6$ and $\Mat{p}_7$) that provide action information from other action instances in the same video but also two proposals ($\Mat{p}_8$ and $\Mat{p}_9$) with background scenes related to the action instance. Lastly, the temporal boundary of $\Mat{p}_1$ is refined to match the corresponding ground truth and the action category is correctly predicted by our method. Clearly, our proposed GCM is able to exploit contextual information to improve the action localization performance.
	
			\begin{table}[!tb]
	\centering
	\caption{Comparison between our model and the MLP baseline on THUMOS14, measured by mAP (\%) when tIoU=0.5..}
	\begin{tabular}{l|cc|cc}
		\hline
		mAP@tIoU=0.5                & RGB     & Gain      & Flow     & Gain        \\ \hline
		MLP$_1$ + MLP$_2$        & 36.82      & -       & 46.74  & -\\
		MLP$_1$ + GCM$_2$  & 38.11   & 1.29       & 47.39 & 0.65\\
		GCM$_1$ + MLP$_2$  & 37.87   &  1.05       & 48.14  & 1.40 \\
		GCM$_1$ + GCM$_2$   & \textbf{39.38}   & \textbf{2.56}       & \textbf{48.76}  &  \textbf{2.02}\\ \hline
	\end{tabular}
	\label{Tab:twographs}
\end{table} 

	\begin{table}[!t]
	\centering
	\caption{Comparison with MLP baseline in terms of runtime, computation complexity in FLOPs, and action localization mAP on THUMOS14. }
	\begin{tabular}{l|c|c|cc}
		\hline
		\multirow{2}{*}{Method} & \multirow{2}{*}{Runtime} & \multirow{2}{*}{FLOPs} & \multicolumn{2}{c}{mAP@tIoU=0.5} \\ \cline{4-5} 
		&                          &                        & RGB        & Flow       \\ \hline
		MLP$_1$ + MLP$_2$         & 0.376s                  & 16.57M                 & 36.82       & 46.74       \\
		GCM$_1$ + GCM$_2$                   & 0.404s                  & 17.70M                 & 39.38       & 48.76       \\ \hline
	\end{tabular}
	\label{tab:runtime}
\end{table}	

		\begin{table}[!tb]
		\centering
		\caption{Comparison between our model and mean-pooling (MP) on THUMOS14, measured by mAP (\%) when tIoU=0.5.}
		\begin{tabular}{l|cc|cc}
			\hline
			mAP@tIoU=0.5                & RGB     & Gain      & Flow     & Gain        \\ \hline
			MP$_1$ + MP$_2$        & 37.12      & -       & 46.96  & -\\
			MP$_1$ + GCM$_2$  & 38.32   & 1.20       & 47.66 & 0.80 \\
			GCM$_1$ + MP$_2$  & 38.38   & 1.26        & 47.93  & 1.07 \\
			GCM$_1$ + GCM$_2$   & \textbf{39.38}   & \textbf{2.26}       & \textbf{48.76}  &  \textbf{1.80}\\ \hline
		\end{tabular}
		\label{Tab:mean-pooling}
	\end{table}

				\begin{figure}[!tb]
		\centering
		\includegraphics[width=\linewidth]{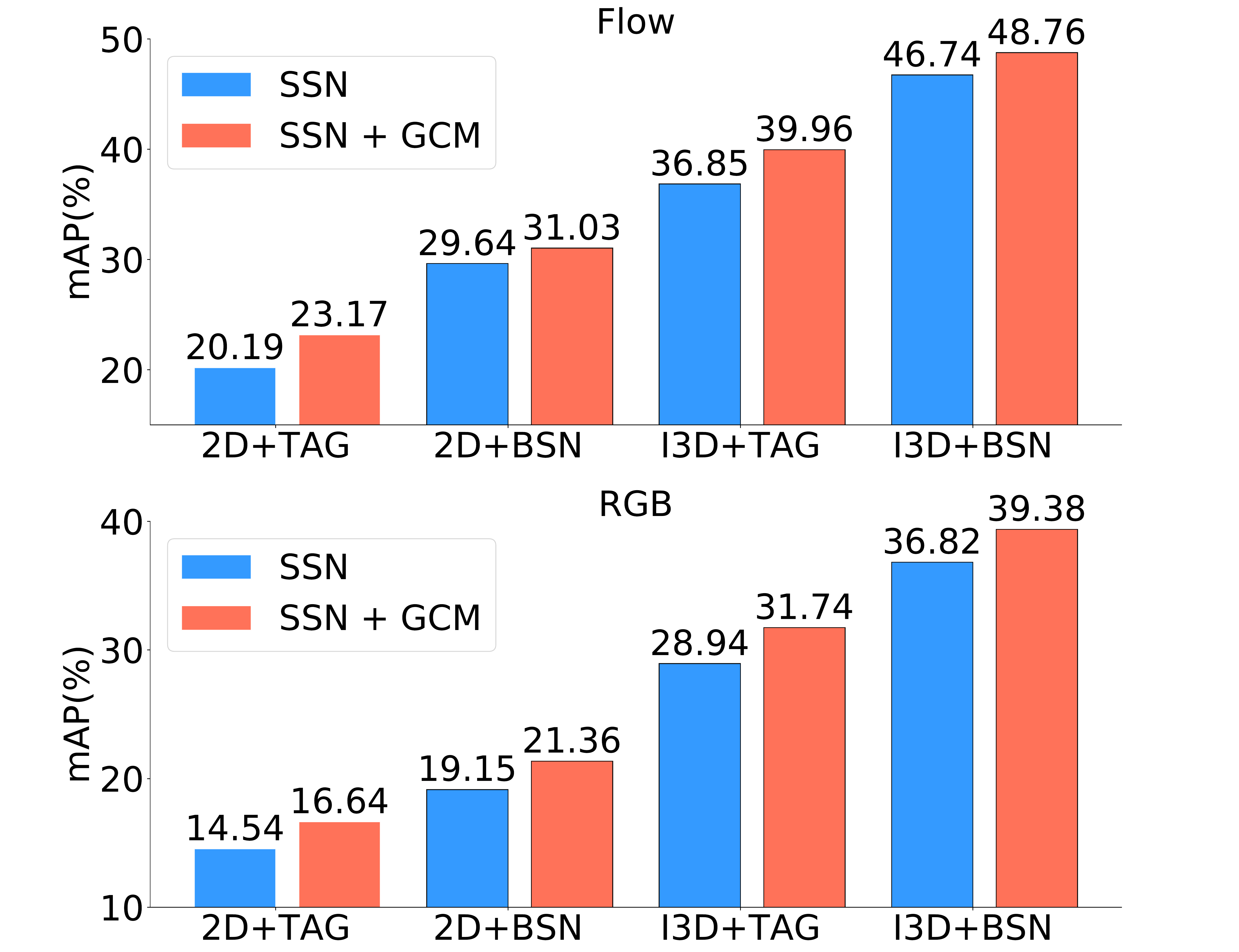}
		\caption{Action localization results on THUMOS14 with different backbones, measured by mAP@tIoU=0.5.}
		\label{Fig:instantiations}
	\end{figure}

	\subsection{How does the graph convolution help?}
	In addition to graph convolutions, performing mean pooling among proposal features is another way to enable information dissemination between action units. We thus conduct another baseline by first adopting MLP on the action unit  features and then conducting mean pooling on the output of MLP over adjacent action units. The adjacent connections are formulated by using the same graph as GCN. We term this baseline as MP below.
	Similar to the setting in Section~\ref{Sec:mlp}, we have three variants of our model including: (1) \textbf{MP$_1$ + MP$_2$}; (2) \textbf{MP$_1$ + GCM$_2$}; and (3) \textbf{GCM$_1$ + MP$_2$}. We report the results in Table \ref{Tab:mean-pooling}. The models with two GCMs outperform all MP variants, demonstrating the superiority of graph convolution over mean pooling in capturing between-proposal connections.
	The protocol \textbf{MP$_1$ + MP$_2$} in Table~\ref{Tab:mean-pooling} performs better than \textbf{MLP$_1$ + MLP$_2$} in Table~\ref{Tab:twographs}, which again reveals the benefit of modeling the  relations between action units, even though we pursue it using the naive mean pooling.
	

	\subsection{Influences of different backbones}\label{Sec:backbone}
	Our framework is general and compatible with different backbones (\ie, action units and features). In addition to the backbones applied above, we further perform experiments on TAG action units~\cite{zhao2017temporal} and 2D features~\cite{lin2018bsn}.
	We try different combinations: (1) BSN+I3D, (2) BSN+2D, (3) TAG+I3D, and (4) TAG+2D, and report the results of SSN and SSN+GCM in Figure \ref{Fig:instantiations}. In comparison with MLP, our method leads to significant and consistent improvements in all types of features and action units. These results conclude that our method is generally effective and is not limited to the specific feature or proposal type. 
			\begin{figure*}[!t]
		\centering
		\includegraphics[width=\linewidth]{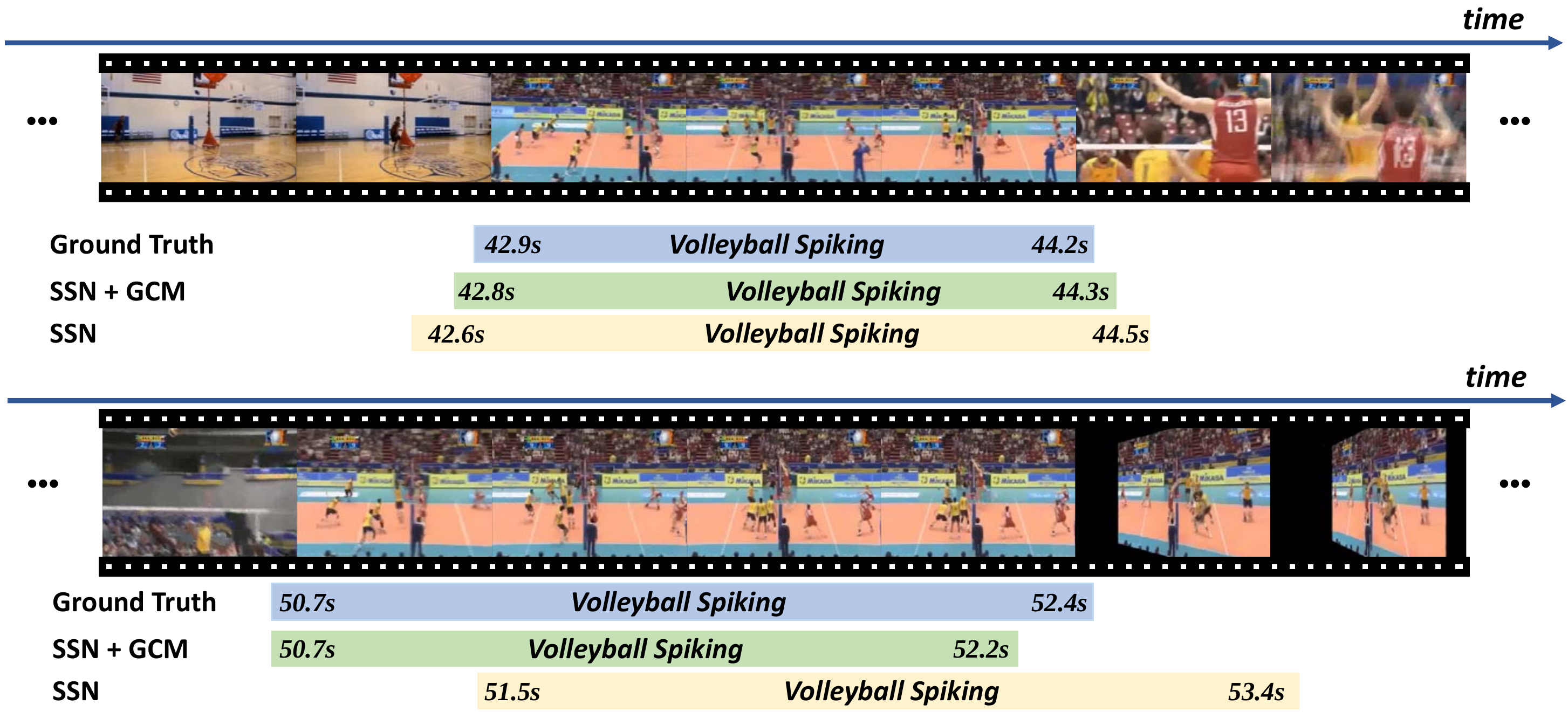}
		\caption{Qualitative results on THUMOS14 dataset. Our proposed GCM helps SSN to predict a more precise temporal boundary. }
		\label{Fig:qualitative}
	\end{figure*}
	\subsection{The weights of edge and self-addition}
	
	We have defined the weights of edges in Eq.~\eqref{Eq:adj}, where the cosine similarity (cos-sim) is applied. This similarity can be further extended by first embedding the features before the cosine computation. We call the embedded version as embed-cos-sim, and compare it with cos-sim in Table \ref{Tab:edge}. 
	No obvious improvement is attained by replacing cos-sim with embed-cos-sim (the mAP difference between them is less than $0.3\%$). Eq.~\eqref{Eq:gcn-sampling} has considered the self-addition of the node feature. We also investigate the importance of this term in Table~\ref{Tab:edge}. It suggests that the self-addition leads to at least 1.06\% absolute improvements on both RGB and flow streams.
			\begin{table}[!tb]
		\caption{Comparison of different types of edge functions on THUMOS14, measured by mAP (\%) when tIoU=0.5.}
		\centering
		\begin{tabular}{l|cc}
			\hline
			mAP@tIoU=0.5                & RGB      & Flow      \\ \hline
			cos-sim   & 38.32    & 47.62   \\
			cos-sim + self-add    & 39.38  & 48.76 \\ 
			embed-cos-sim + self-add   & 39.27   & 48.92 \\ \hline 
		\end{tabular}
		\label{Tab:edge}
	\end{table}

	\noindent \textbf{Comparisons with learned weights.} To further verify the effectiveness of our graph construction strategy, we conduct an experiment by using learned weights of edge. Specifically, we first construct a fully-connected graph and then follow the ``scaled dot-product attention'' mechanism in~\cite{vaswani2017attention} to obtain the adjacent matrix by computing
        	$\bA_{ij} = \frac{e^{(\bW_1\bx_i)^T  (\bW_2\bx_j)}}{\sum_{n=1}^{N} e^{(\bW_1\bx_i)^T  (\bW_2\bx_n)}}$,
	where $\bW_1$ and $\bW_2$ are learnable parameters and $N$ is the number of proposals in one video. 
	Note that one video often contains thousands of proposals, and thus using a fully-connected graph will inevitably incur large computation cost when aggregating information from all other proposals. From Table~\ref{Tab:self-att}, our GCM outperforms the baseline using learned weights. This is probably because the fully-connected graph may introduce noise from irrelevant proposals, which may even make the training unstable. In contrast, our GCM passes messages only from the temporally adjacent and semantically correlated action units, and thus may eliminate noisy information from irrelevant action units and yield better performance. Moreover, using learned weights is able to lift the action localization performance of the baseline (49.7\% vs 49.3\%). These results reveal that exploiting action unit relations helps localize actions more precisely, and they also justify our motivation for considering the relations between action units.

	    
	    \begin{table}[!t]
	    
		\centering
		\caption{Comparisons between our GCM and the baseline using learned weights on THUMOS14.}
		\begin{tabular}{l|ccccl}
			\hline
			\multirow{2}{*}{Method}   & \multicolumn{5}{c}{mAP at different tIoUs} \\  \cline{2-6}
			            & 0.1    & 0.2  & 0.3 & 0.4  &0.5       \\ \hline
			SSN~\cite{zhao2017temporal}    & 69.7 & 67.5 & 64.6 & 58.3 & 49.3       \\ 
			SSN + Learned weights  & 70.3 & 68.3 & 64.5 & 58.2 & 49.7 ($\uparrow$0.4)       \\
			SSN + GCM (ours) & \textbf{70.5} & \textbf{68.6} & \textbf{65.2} & \textbf{59.8}  & \textbf{50.9 ($\uparrow$1.6)}        \\ \hline
		\end{tabular}
		\label{Tab:self-att}
	\end{table}
	
	\subsection{Is it necessary to consider three types of edges?}
	
	To evaluate the necessity of formulating three types of edges, we perform experiments on three variants of our method, each of which removes one type of edge in the graph construction stage. From Table~\ref{Tab:surrounding}, the result drops remarkably when any kind of edge is removed. 
	Another crucial point is that our method still improves the MLP baseline when only the surrounding edges remain. The rationale behind this could be that actions in the same video are correlated and exploiting the surrounding relation enables more accurate action classification. 
	
	\begin{table}[!tb]
		\centering
		\caption{Comparison of three types of edge on THUMOS14, measured by mAP (\%) when tIoU=0.5.}
		\begin{tabular}{l|cc|cc}
			\hline
			mAP@tIoU=0.5                & RGB     & Gain      & Flow     & Gain        \\ \hline
			w/ all edges   & 39.38   & -       & 48.76  &  - \\
			w/o surrounding edges   & 38.80   & -0.58      & 47.69  & -0.56 \\ 
			w/o contextual edges   & 38.28   & -1.10      & 47.57   & -0.68 \\ 
			w/o semantic edges   & 39.02  & -0.36      & 47.38  & -0.87 \\ 
			w/o  edges (MLP)    & 36.82   &  -2.56    &  46.74     & -2.02 \\ \hline
		\end{tabular}
		\label{Tab:surrounding}
	\end{table}

	\begin{table}[!t]
	\centering
	\tabcolsep 5pt 
	\caption{Comparison of different sampling sizes and training time for each iteration on THUMOS14, measured by mAP@tIoU=0.5.}
	\vspace{0.1cm}
	\begin{tabular}{c|cccccc}
		\hline
		$N_s$  & 1 & 2 & 3 & 4 & 5 & 10 \\ \hline
		mAP & 48.28 & 48.47 & 48.54 & \textbf{48.76} & 48.34 & 48.30\\
		Time(s) & 0.10 & 0.23 & 0.33 & 0.41 & 0.48 & 1.72 \\ \hline
	\end{tabular}
	\label{Tab:sampling}
\end{table}

\begin{table}[!tb]
	\centering
	\caption{Ablation study of our GCM on D-SSAD, measured by mAP (\%) when tIoU=0.5 on THUMOS14.}
	\begin{tabular}{l|cc}
		\hline
		Setting                & mAP@IoU=0.5     & Gain            \\ \hline
		D-SSAD~\cite{huang2019decoupling} (our impl.)       & 43.21      & -\\
		D-SSAD + GCM $\times$ 1   & 43.47   & 0.26       \\
		D-SSAD  + GCM $\times$ 2  & 44.29   & 1.08      \\
		D-SSAD + GCM $\times$ 3   & \textbf{44.77}  & 1.56 \\ \hline
	\end{tabular}
	\label{Tab:module}
\end{table} 
	
	\subsection{The efficiency of our sampling strategy}
	
	We train our model efficiently based on the neighborhood sampling in Eq.~\eqref{Eq:gcn-sampling}. Here, we are interested in how the sampling size $N_{s}$ affects the final performance. Table \ref{Tab:sampling} reports the testing mAPs corresponding to different $N_s$ values varying from 1 to 5 (and also 10). The training time per iteration is also added in Table \ref{Tab:sampling}. 
	We observe that when $N_{s}=4$, the model achieves higher mAP than the full model (\ie, $N_{s}=10$) while reducing  the training time by 76\% for each iteration. This is interesting, as sampling fewer nodes yields even better results. We conjecture that the neighborhood sampling can bring in more stochasticity and guide our model to escape from the local minimal during training, thus delivering better results.

	
	\begin{figure*}[!t]
	\centering
	\includegraphics[width=\linewidth]{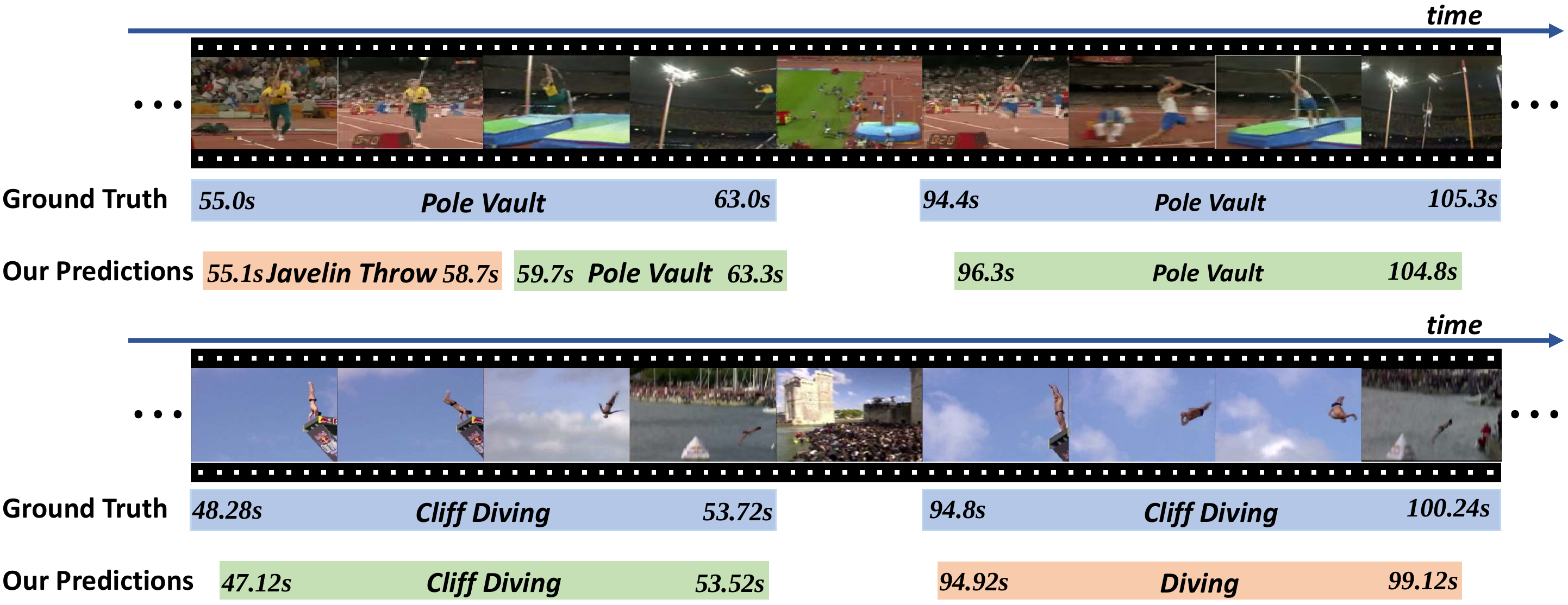}
    \caption{Examples of failure cases. \textbf{Top:} Our method predicts the beginning portion of \emph{Pole Vault} as \emph{Javelin Throw} since these two actions have similar contents (\ie, an athlete running with a pole).
    \textbf{Bottom:} Our method mis-classifies the action \emph{Cliff Diving} into the action \emph{Diving} without recognizing the background \emph{cliff}.
    }
	\label{Fig:failure}
    \end{figure*}

	\section{Ablation Results on One-Stage Methods}

	
	\subsection{Effectiveness of GCM}
	\noindent \textbf{Decoupled single-shot temporal action detection (D-SSAD)~\cite{huang2019decoupling}}. \hao{Huang \etal decoupled the localization and classification in a one-stage scheme.
		In particular, D-SSAD consists of three main components: a base feature network, an anchor network, and a classification/regression module. The base feature network extracts representations of each video segment to form feature maps. Then, a multi-branch anchor network takes the feature maps as input and produces multiple anchors at each location on the feature maps. Last, the anchors are processed by the classification and regression module. In our experiments, we add our GCM to the feature maps before generating anchors. As discussed in Section~\ref{Sec:one-stage}, the GCM can be inserted one or multiple times throughout the network to model the feature relationships at different scales. Therefore, we add the GCM to feature maps with multiple scales (from 1 to 3). From Table~\ref{Tab:module}, the performance of D-SSAD is improved by using our GCM to enhance the features. As more GCMs are inserted, the action localization results increase, which demonstrates that our GCM is general and compatible with the one-stage action localization methods. }
	

		\begin{table}[!tb]
		\centering
		\caption{Comparison of two types of edge on THUMOS14, conducted on D-SSAD~\cite{huang2019decoupling} with GCM.}
		\begin{tabular}{cc|c|c}
			\hline
			\multicolumn{2}{c|}{Settings}       & \multirow{2}{*}{mAP@tIoU=0.5} & \multirow{2}{*}{Gain} \\ \cline{1-2}
			surrounding edges & semantic edges &                               &                       \\ \hline
			$\times$   &    $\times$        &  43.21    & -   \\
\checkmark   &    $\times$        &  43.82    & 0.61   \\
$\times$ &       \checkmark            &   44.19    & 0.98  \\
\checkmark        &       \checkmark           &   \textbf{44.77}   & 1.56   \\
 \hline
		\end{tabular}
%
		\label{Tab:dssad-edge}
	\end{table}

	\subsection{How much does each type of edge help?}
	
	To evaluate the effectiveness of surrounding and semantic edges, we perform experiments by gradually adding one type of edge to our GCM.
	From Table~\ref{Tab:dssad-edge}, adding the surrounding edge and semantic edge to the baseline (\ie, without both types of edges) results in at least 0.61\% improvements in terms of action localization mAP. When considering both surrounding and semantic edges simultaneously, the performance is further improved to 44.77\%, which strongly supports the necessity of constructing two types of edges in our proposed GCM.

	%

	\section{Qualitative Results}
	
	Given the significant improvements, we also attempt to find out in what cases our method improves over the baseline method. We visualize the qualitative results on THUMOS14 in Figure~\ref{Fig:qualitative}. In these examples, the baseline method (\ie, SSN~\cite{zhao2017temporal})
	is able to predict the action category correctly, while failing to precisely predict the location of actions. With the help of our proposed GCM, we predict a more precise temporal boundary, which demonstrates the effectiveness of GCM for temporal action localization. 
	
	\noindent \textbf{Failure case analysis.} 
	Our method achieves state-of-the-art performance on
    two benchmark action localization datasets, but like other methods, it is still not sufficiently capable of detecting actions when they share the similar contents.
    For example, in Figure~\ref{Fig:failure}, our method correctly detects the locations of actions but misclassifies the action \emph{Pole Vault} into the action \emph{Javelin Throw} since both these actions share similar contents (\ie an athlete runs when holding a pole). Another failure case is the misclassification between \emph{Cliff Diving} and \emph{Diving}. While this is a common challenge in temporal action localization, exploiting more advanced feature extraction methods may solve it to some extent, which will be left for future exploration.
    
	

	\section{Conclusions}\label{Sec:conclusion}
	
	In this paper, we have exploited the relationships between action units to address the task of temporal action localization in videos. Specifically, we have proposed to construct a graph of action units based on the temporal context and semantic information, and apply GCNs to enable message passing among action units. In this way, we enhanced the action unit features and eventually improved the action localization performance. More critically, we have integrated the above graph construction and graph convolution processes into a general graph convolutional module (GCM), which can be easily inserted into existing action localization methods, including the one-stage paradigm and two-stage paradigm. Experimental results show that our GCM is compatible  with other action localization methods and helps to consistently improve their action localization accuracy. With the aid of GCM, our method outperforms the state-of-the-art methods by a large margin on two benchmarks, \ie, THUMOS14 and ActivithNet v1.3. 
	It would be interesting to extend our method for object detection in images and we leave it for our future work.

	\section*{Acknowledgments}
	This work was partially supported by the Scientific Innovation 2030 Major Project for New Generation of AI under Grant 2020AAA0107300, Ministry of Science and Technology of the People's Republic of China,
	National Natural Science Foundation of China (NSFC) 62072190,
	Key-Area Research and Development Program of Guangdong Province 2018B010107001, Program for Guangdong Introducing Innovative and Enterpreneurial Teams 2017ZT07X183. This work was jointly sponsored by CAAI MindSpore Open Fund. Runhao Zeng and Wenbing Huang contributed equally to this work.
	
	
	


	\ifCLASSOPTIONcaptionsoff
	\newpage
	\fi

	
	
	%
	%
	%

	\bibliographystyle{abbrv}
	\bibliography{egbib}

\end{document}